%% file: root.tex
\documentclass[journal]{IEEEtran}
\usepackage{cite}
\usepackage{amsmath,amssymb,amsfonts}

\usepackage{algorithm}
\usepackage{algorithmic}
\usepackage{setspace}
\usepackage{mathrsfs}

\usepackage{graphicx}
\usepackage{textcomp}

\usepackage{picinpar}
\usepackage{flushend}
\usepackage[latin1]{inputenc}
\usepackage{soul}
\usepackage{multirow}
\usepackage{pifont}
\usepackage{alltt}
\usepackage[hidelinks]{hyperref}
\usepackage{enumerate}
\usepackage{siunitx}
\usepackage{epstopdf}
\usepackage{pbox}
\usepackage{mathtools}
\usepackage{latexsym}
\usepackage{bm}
\usepackage{multicol}
\usepackage{pifont}
\usepackage{lipsum}
\usepackage{tabularx}
\usepackage{array}
\usepackage{float}
\usepackage{booktabs}
\usepackage{multirow}
\usepackage{tikz}
\usepackage{upgreek}
\usepackage{makecell}
\usepackage{threeparttable}
\usepackage{sidecap}

\usepackage{soul}
\soulregister\cite7
\soulregister\ref7 
\soulregister\citep7
\soulregister\citet7
\soulregister\pageref7

\def\BibTeX{{\rm B\kern-.05em{\sc i\kern-.025em b}\kern-.08em
    T\kern-.1667em\lower.7ex\hbox{E}\kern-.125emX}}

\begin{document}
\title{Nav-SCOPE: Swarm Robot Cooperative Perception and Coordinated Navigation}

\author{
	\vskip 1em
	Chenxi Li, \emph{Student Member, IEEE},
	  Weining Lu,
        Qingquan Lin, \\
        Litong Meng, \emph{Member, IEEE}, Haolu Li,
	Bin Liang, \emph{Senior Member, IEEE}
        \vskip 1em

\thanks{This work is supported by the BNRist project under Grants BNR2024TD03003, and the National Natural Science Foundation of China under Grants 92248304. \textit{(Corresponding author: Weining Lu.)}}
\thanks{Chenxi Li, Qingquan Lin, Litong Meng and Bin Liang are with the Department of Automation, Tsinghua University, Beijing, 100084, China. (e-mail: \tt\small lcx22@mails.tsinghua.edu.cn, linqq19@tsinghua.org.cn, menglt@ieee.org, liangbin@tsinghua.edu.cn).}%
\thanks{Weining Lu is with the Beijing National Research Center for Information Science and Technology, Tsinghua University, Beijing, 100084, China. (e-mail: \tt\small luwn@tsinghua.edu.cn).}%
\thanks{Haolu Li is with the School of Mechanical and Electrical Engineering, Beijing Information Science and Technology University, Beijing, 100192, China. (e-mail: \tt\small 202402005@bistu.edu.cn).}%
}

\maketitle

\begin{abstract}
This paper proposes a lightweight decentralized solution for multi-robot coordinated navigation with cooperative perception. First, we introduce a rapid way to process sensory data, thus obtaining safe directions and key environmental features. Then, an information flow is created to facilitate real-time perception sharing over wireless ad-hoc networks. Consequently, the environmental uncertainties of each robot are reduced by interaction fields that deliver complementary information. Finally, path optimization is achieved in a probabilistic way, enabling self-organized coordination with effective convergence, divergence, and collision avoidance. Our method is fully interpretable and ready for deployment without gaps. Comprehensive simulations and real-world experiments demonstrate reduced path redundancy, robust performance across various tasks, and minimal demands on computation and communication.
\end{abstract}

\def\abstractname{Note to Practitioners}
\begin{abstract}
Local perception is the information source for robot navigation in unknown environments. The extended perception at the swarm level provides complementary information for each robot. This can optimize robot paths and achieve coordinated navigation in a distributed way. However, this purpose is currently unachievable due to the high demands of communication and computation on board. This study reduces these two demands to the extreme by Fast Fourier Transform (FFT), digital filtering, probabilistic fusion, and swarm interaction forces. At the same time, our fully interpretable method avoids sim-to-real gaps and can be deployed directly on ground mobile robots. Simulation and real-world experiments outperform a state-of-the-art approach. Future work aims to extend our method to aerial robots as drones.
\end{abstract}

\begin{IEEEkeywords}
Multiple mobile robots, cooperative systems, distributed robot systems
\end{IEEEkeywords}

\input{Introduction}
\input{Related_Work}
\input{Preliminaries}
\input{Methods}
\input{Simulation_Results}
\input{Real_World_Experiments}

\input{Conclusion}

\bibliographystyle{IEEEtran}
\bibliography{IEEEabrv,egbib.bib}

\end{document}

%% file: Introduction.tex
\section{Introduction}
\label{sec:Introduction}
In recent years, multi-robot systems have shown great potential in wild \cite{Swarm2022}, industry \cite{Standalone2021}, and indoor services \cite{Minimal2019}. Self-organized swarms stand out for their better automation and wider application.
Despite significant progress, most solutions only use each robot's local perception independently for navigation, while few studies have so far utilized real-time cooperative perception of the swarm to optimize the planning process. Consequently, robots' coordination is often confined to leader-follower \cite{Modular2022,Guidance2024} or collision avoidance \cite{Swarm2022,STR2025} due to limited information flow within swarm systems, in which robots often encounter unexpected obstructions or unnecessary detours.

The advantages of cooperative perception are obvious: a wider field of view, more details of the unknown environment, and that lead to better paths with fusion optimization. The challenges include but are not limited to informative environmental representation, decision-making with fused perception, on-board processing latency, and minimal unreliable communication. The first two are method requirements, and the latter two are hardware constraints.

\begin{figure}[t]
  \centering
  \includegraphics[width=\linewidth,keepaspectratio]{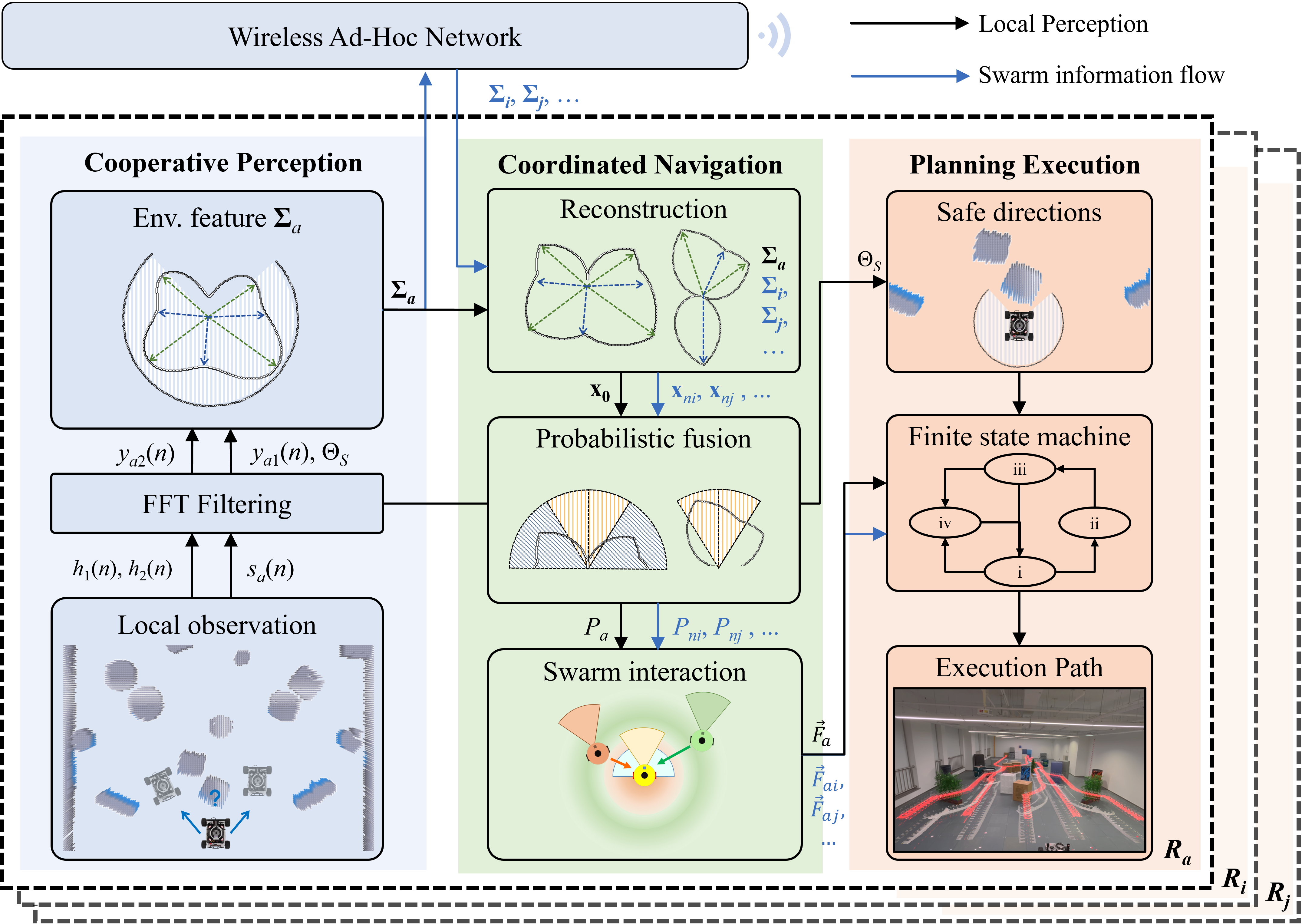}
  \vspace{-0.5cm}
  \caption{System architecture.}
  \label{fig:intro}
  \vspace{-0.5cm}
\end{figure}

This paper proposes Nav-SCOPE: \textbf{S}warm \textbf{CO}operative \textbf{PE}rception and coordinated \textbf{Nav}igation. This is an extremely lightweight systematic solution that addresses the above contradiction.
In this system, each robot is used as a telescope of neighboring robots for an extended field of view. When robots decide their future paths, they will use the fused perception to choose better directions with fewer detours or conflicts. Experimental results show self-organized paths and better swarm automation. To the best of our knowledge, Nav-SCOPE is the first interpretable navigation solution with real-time cooperative perception, and it does not require a mapping process or external localization. The paper makes the following contributions:

\begin{itemize}
    \item The formulation of perception and fusion is established systematically and interpretively with minimal computation and communication requirements.
    
    \item The navigation process is detailed for better swarm coordination and less path redundancy. Parameters can be determined directly before deployment.

    \item Comprehensive simulations in challenging environments demonstrate that Nav-SCOPE outperforms a previously state-of-the-art approach. 
    Real-world experiments show that our method can be applied robustly without gaps.
\end{itemize}

This article is organized as follows. Section \ref{sec:related work} presents the related work. Section \ref{sec:preliminaries} gives the preliminaries of our work. In Section \ref{sec:formulation of Nav-SCOPE}, the formulation of Nav-SCOPE is detailed, including sensor data processing, information flow from cooperative perception, coordinated navigation, and path optimization. The necessary steps for the deployment of Nav-SCOPE are introduced in Section \ref{sec:parameter determination and model deployment}. Simulation and real word experiments are presented in Section \ref{sec:simulation results} and Section \ref{sec:real world experiments}, respectively. Finally, Section \ref{sec:Conclusion} presents the conclusions of this study.

%% file: Related_Work.tex
\section{Related Works}
\label{sec:related work}
Numerous studies have focused on local planning since the advent of mobile robots. Autonomous navigation was first realized in individual robots. Reactive strategies, such as Potential Fields (PFs) \cite{Realtime1985, Predictive2021} and Bugs \cite{Dynamic1986, new2012}, are direct and lightweight solutions. Planners based on voxel maps \cite{EGO-Planner2021, UFOExplorer2022} can provide better local trajectories with more environmental details. Learning-based methods such as reinforcement learning \cite{Champion2023} and guidance planning \cite{YOPO2024} have also excelled for their automation and become popular in recent years. These methods do not require a pre-built map, as a Simultaneous Localization and Mapping (SLAM) \cite{FastSLAM2002} process before planning execution is not suitable for many tasks and cases.

As robot systems grow to fleets and swarms, critical issues of coordination between robots need to be solved. Centralized control is a typical way to regulate robot paths. For example, \cite{Predictive2021} and \cite{Collaborative2025} used central computing stations to achieve flock flight on multiple quadrotors. The former is a typical solution based on optimization, while the latter is achieved by reinforcement learning.
However, these approaches require ideal communication and external localization, making them vulnerable in scaled-up scenarios \cite{RACER2023}. \cite{Minimal2019, Standalone2021} used received signal strength indication (RSSI) to achieve distributed navigation, but still needed external beacons to provide a priori environmental information, which limits the scalability of the algorithm.

Decentralized systems can be more versatile in large unknown environments. The more information utilized, the better performance a system would have. Currently, it is common to use mutual positioning and trajectory sharing to achieve self-organization and swarm navigation. Consequently, recent studies have provided two types of solutions. (i) Leader-follower structures achieve synergic formation towards the same goal, such as leading robot following \mbox{\cite{Formation2015}}, formation control \mbox{\cite{Modular2022, Guidance2024}} and target tracking \mbox{\cite{DistributedEstimation2022, Consensus2022}}, which are mainly based on distance keeping and barycentric coordinates. (ii) Collision avoidance structures optimize robot paths to reduce negative interactions in various tasks, such as spatial-temporal optimization of trajectories \mbox{\cite{EGO-Swarm2021,Swarm2022}}, and similar approaches performed by neural networks \mbox{\cite{Reciprocal2024,STR2025}}. A typical state-of-the-art (SOTA) solution is EGO-Swarm-v2 \mbox{\cite{Swarm2022}}, which was tested in wild bamboo forests.

Cooperative perception offers external environmental information to each robot, thereby holding significant potential to enhance swarm navigation and improve their performance. However, it is challenging to form an information flow among robots, as well as to take advantage of it.
\mbox{\cite{Multi-Robot2022, Graph2024}} proposed Graph Neural Network (GNN) pipelines to share perceptual features, while the swarm planning process was not yet considered. \mbox{\cite{CollaborativeN2025}} achieved a systematic GNN framework from perception to planning execution, but robot actions were limited to discrete on grid maps, thereby suffering from sim-to-real gaps. In contrast, Nav-SCOPE is designed to solve these challenges and can be immediately deployed in real-world applications.

%% file: Preliminaries.tex
\section{Preliminaries}
\label{sec:preliminaries}
\begin{SCfigure}[0.5][th]
  \includegraphics[width=0.5\linewidth,keepaspectratio]{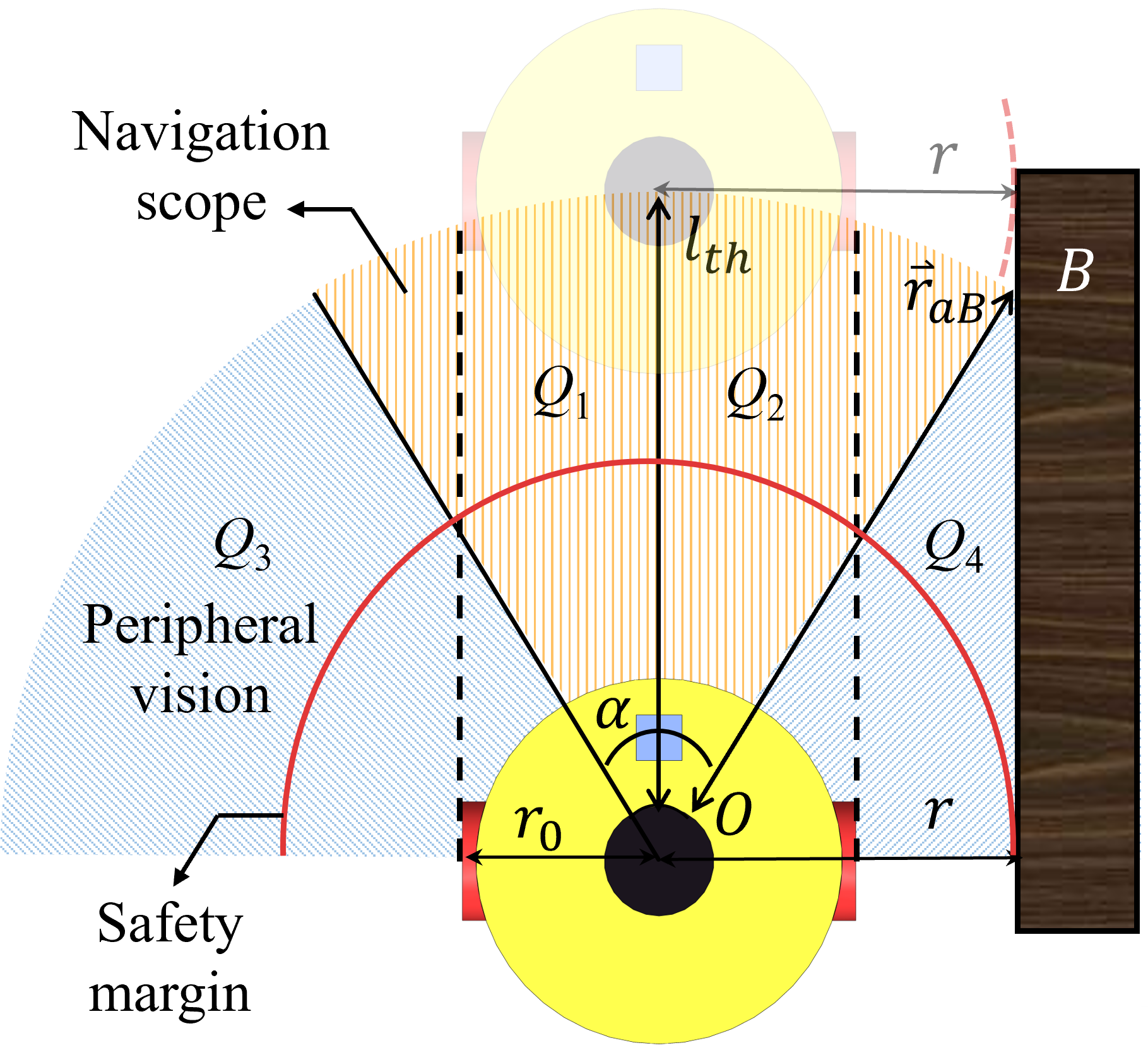}
  \vspace{-0.4cm}
  \caption{Safe margin and four observation quadrants. If there is no obstacle within the orange scope, the robot can keep a safety margin $r$ from the obstacle $B$.}
  \label{fig:safe_margin}
\end{SCfigure}

We consider a swarm of ground mobile robots $R_i, i\in\{1,2,\dots,N\}$ operating in unknown cluttered environments. Each robot has its local perception within the maximum range of $R$. The sensory data is defined as a digital depth sequence $s_i(n)$, with $M$ uniform samples within its field of view (FOV):
\begin{equation}
s_i(n) = \mathbf{d}\left[\theta_n \in \Theta \right] \leq R, n=1,2,\dots,M,
\label{eq:sensor data}
\end{equation}
where $\Theta$ is the FOV, and $\mathbf{d}$ is the sequent depth array obtained from LiDARs or cameras, with each item corresponding to a distance in a certain direction $\theta_n$. Consider the physical dimension $r_0$ of a robot, a safety margin $r$ is used to keep the robot from obstacles, as shown in Fig. \ref{fig:safe_margin}. The FOV is divided into four quadrants: navigation scope ($Q_{1}$, $Q_{2}$), and peripheral vision ($Q_{3}$, $Q_{4}$). They are used to evaluate the occupation of the future route. The scope angle \( \alpha \) and planning distance \( l_{th} \) are determined by geometric relationships:
\begin{equation}
\alpha = \pi - 2\arccos{\left( \frac{r_0}{r}\right)}, l_{th} = \frac{r}{\cos{\left( \frac{\pi - \alpha}{2}\right)}} = \frac{r^2}{r_0}.
\label{eq:alpha}
\end{equation}
The necessary and sufficient condition of a safe direction can be expressed as a safety window in $Q_{1}$ and $Q_{2}$:
\begin{multline}
    \theta_m \text{ is safe} \iff \text{for } n \in (m-\frac{\alpha}{2}, m+\frac{\alpha}{2}),
    s_i(n) \geq l_{th}.
\label{eq:necessary and sufficient condition for safe}
\end{multline}

In this study, each robot can advance or rotate left and right towards a safe direction. The task of the swarm is to navigate safely and coordinately towards targets. The objective of Nav-SCOPE is to provide forward directions to robots, thus minimizing swarm path lengths with the least collision and resource consumption.

%% file: Methods.tex
\section{Formulation of Nav-SCOPE}
\label{sec:formulation of Nav-SCOPE}
The system architecture of Nav-SCOPE is illustrated in Fig. \ref{fig:intro}. In this section, we will first introduce the FFT process and our filter design. Then, the information flow among swarm robots is formed. Finally, coordinated navigation facilitated by cooperative perception is elaborated.
\begin{figure*}[t]
  \centering
  \includegraphics[width=\linewidth,keepaspectratio]{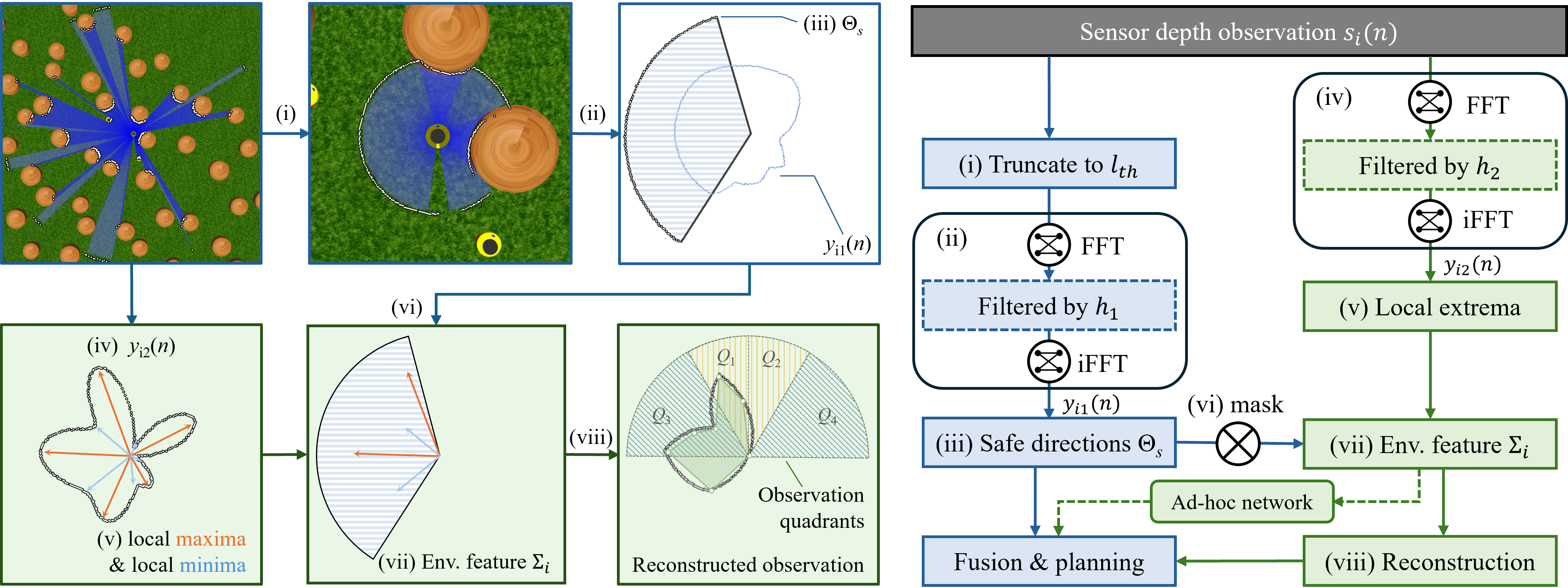}
  \vspace{-0.5cm}
  \caption{Sensor data processing framework (right) and an illustrative example (left) in local coordinate. Blue boxes identify safe directions and green boxes extract environmental features. The feature excludes narrow spaces beneath the robot dimension, while maintains prospective directions  for reconstruction. 
  The computation complexity in (ii) and (iv) is $O(m \log m)$, while others are less than $O(m)$. The volume of environmental feature $\Sigma_i$ is usually less than $(2\Theta/\alpha)$.
  }
  \label{fig:filtering_process}
  \vspace{-0.4cm}
\end{figure*}
\subsection{FFT Process and Filter Design}
\label{subsec:sensor data processing}
In the perception stage, Nav-SCOPE scans to evaluate all directions of $s_i(n)$ within the FOV. The evaluation process can be performed rapidly by digital filtering based on Fast Fourier Transform (FFT), which was demonstrated in a previous study \mbox{\cite{li2024highly}}, since convolution in the time domain is equivariant to multiplication in the frequency domain. The overall complexity is $O(m\log m)$ in computation, and an illustration of the framework is shown in Fig. \ref{fig:filtering_process} (i)-(iv).
The evaluation of this stage serves two functions: first, to identify safe routes and second, to extract key features of the surrounding environment.

The design of digital filters is critical for FFT filtering. In order to achieve the two functions discussed above, we propose two types of filters, respectively.
First, to identify safe routes subject to robot dimensions $r_0$ and planning distance $l_{th}$, a normalized rectangular window can be designed to evaluate the occupancy of each direction, which is mathematically equivalent to (\ref{eq:necessary and sufficient condition for safe}). The size of the window $T_c$ corresponds to $\alpha$ in the physical world:
\begin{equation}
h_1(n) = \frac{1}{T_c} R_{T_c}(n - \tau), n=1,2,\dots,M,
\label{eq:h1}
\end{equation}
where $T_c=M (\alpha / \Theta)$ refers to the window size normalized by the digital resolution, and \(\tau = \frac{M - 1}{2}\) is the group delay. This filter operates on short-term perception within the planning distance, so $s_i(n)$ is truncated to $l_{th}$ before being filtered. The safety condition (\ref{eq:necessary and sufficient condition for safe}) is further simplified with the filtered sequence $y_{i1}(n)$:
\begin{equation}
    \theta_m \text{ is safe} \iff y_{i1}(m) \geq l_{th},
\label{eq:y1 for safe}
\end{equation}
and we obtain the set of safe directions $\Theta_s = \{\theta_m|y_{i1}(m) \geq l_{th}\}$ for local planning, e.g., the sector in Fig. \ref{fig:filtering_process} (iii).

Second, Nav-SCOPE is required to evaluate environmental occupancy at the level of perceptual range in the long term. A low-pass filter can effectively extract overall spacial tendency, while filtering out minor details for subsequent compression. Consequently, another filter $h_2$ is designed to extract the key features of open space and potential obstructions according to the same window size $T_c$:
\begin{equation}
h_2(n) = 2f_c \text{sinc} \left(2f_c(n-\tau) \right)w_{M}(n, \beta),
\label{eq:h2}
\end{equation}
where $f_c = 1 / T_c$, $\text{sinc}(x)=sin(\pi x) / (\pi x)$, and $w_{M}(n, \beta)$ is a Kaiser window \cite{Kaiser1980} with parameter $\beta$. As illustrated in Fig. \ref{fig:filtering_process}(iv), the filtered sequence $y_{i2}(n)$ contains several sectors. The centers of these sectors are prospective pathways in long-term planning, and the areas are approximate estimates of open spaces.

\subsection{Information Flow from Cooperative Perception}
\label{subsec:information flow from cooperative perception}
The information flow within the swarm is formed by distributed sharing of robot perception. In order to minimize communication loads, Nav-SCOPE performs informative encoding at transmitter, and rapid reconstruction at receiver. 
The centers and borders of the sectors in $y_{i2}(n)$ indicate possible open spaces and potential obstacles, respectively. Therefore, they are key perceptual features encoded for data transmission. The features can be obtained by calculating the local extrema of $y_{i2}(n)$, which are then masked by safe directions $\Theta_s$ to filter out narrow spaces relative to robot size (Fig. \ref{fig:filtering_process} (vi)-(vii)):
\begin{equation}
\mathbf{\Sigma}_i = \left \{(\phi_{i,j}, d_{i,j}) | \Delta y_{i2}(j) \Delta y_{i2}(j+1) < 0 \cap \Theta_s \right\},
\label{eq:extrema}
\end{equation}
where \(\phi_{i,j}\) and \(d_{i,j}\) are the directions and distances of the extrema. Consequently, we get perceptual features of robots only at the byte level.
At the receiver, the directions \(\phi_{i,j}\) should be calibrated according to the relative pose of two robots. Then, we use simple linear interpolation for approximate reconstruction, as shown in Fig. \ref{fig:filtering_process} (viii), and the information flow within the swarm is established.

In our applications in unknown, cluttered environments, robots are often obstructed by obstacles. The objective of the information flow is to deliver complementary observation of the surrounding environment. Consequently, an effective expansion should keep an overlap between the perceptual regions of two robots. This condition ensures correlation and accessibility between these observations. Hence, robots only connect to the neighboring robots within the FOV, which does not need a strongly connected network, but is time-varying and forms local groups spontaneously. In addition, the information flow of Nav-SCOPE is event-triggered and in an ad hoc way. When a robot needs cooperative perceptions from its neighbors, it will request data transmissions, and then the information flows in a one-way direction to the robot in need. In summary, the transmission mechanism is as follows:
\vspace{0.15cm}
\begin{itemize}
    \item First, cooperative perception activates only among neighboring robots within visual range. This can prevent irrelevant information from robots too far away.
    \item Second, a visual persistence mechanism is introduced to address temporary visual loss towards neighbors.
    \item Third, the observation shares only when needed, such as the time to decide future navigation directions.
\end{itemize}
Thus, the data are minimized while sufficient information is maintained, and the complexity of transmission scales linearly with the number of observed neighbors.
The reduction in communication resource improves reliability under rigorous conditions or unreliable networks.

\begin{figure*}[t]
  \centering
  \includegraphics[width=\linewidth,keepaspectratio]{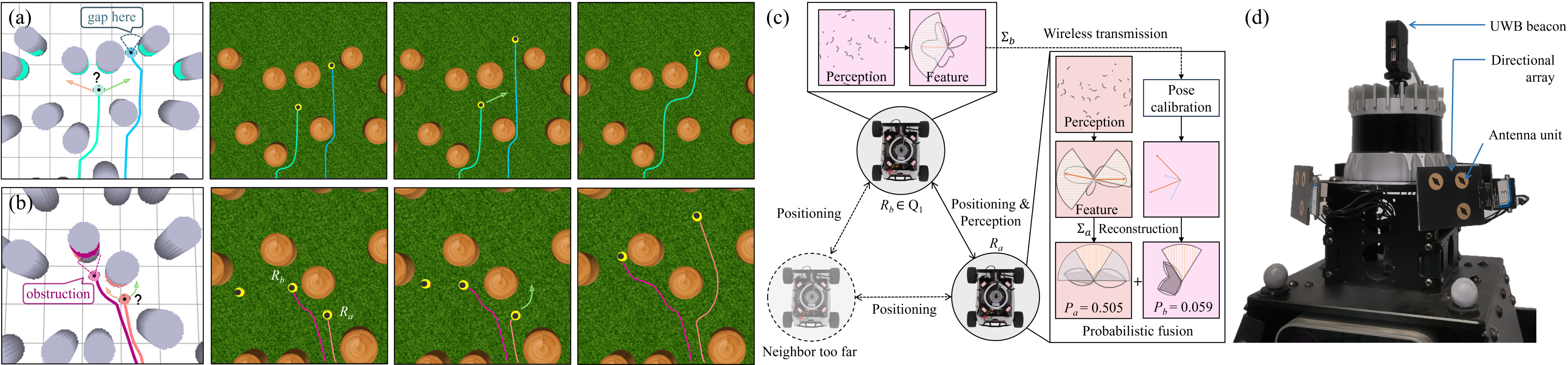}
  \vspace{-0.5cm}
  \caption{Detailed cases of coordination in simulation. (a) Convergence case. (b) Divergence case. (c) A detailed illustration of fusion optimization in (b). $R_a$ has similar probabilities in either directions before fusion. The perception of it's neighbor $R_b$ in quadrant $Q_1$ provides strong probability of obstruction. Consequently, $R_a$ choose to navigate rightwards. (d) Distributed ultra-wide band (UWB) model of each robot for communication and mutual positioning.}
  \label{fig:detail-result}
  \vspace{-0.4cm}
\end{figure*}

\subsection{Coordinated Navigation and Path Optimization}
\label{subsec:Coordinated Navigation}
The objective of coordinated navigation is to optimize robot paths in a distributed way, combining local perception with swarm information flow.
Here, neighboring robots perform as telescopes to provide complementary information in terms of the environment, in other words, help to find open spaces with fewer obstacles towards the target. We use forces to indicate the level of impact.
\subsubsection{Individual intelligence} Robots use their local perception to evoke a tendency in a certain direction within $\Theta_s$. Considering a robot $R_a$, we use a force $\vec{F}_a$ to quantify its motivation. Note that there are two types of turning directions, namely left and right, for ground robots, the amplitude of $\vec{F}_a$ is designed as the probability of searching its path on the left side. If there is no obstruction within the planning distance $l_{th}$, $R_a$ will head towards the target direction $\vec{e}_a$ with the same probability on both sides. Otherwise, when an obstacle is detected in the navigation scope, $R_a$ should select a direction for circumnavigation:
\begin{equation}
\vec{F}_a = P_a \vec{e}_a,
\label{eq:F_a}
\end{equation}
where $P_a$ uses a sigmoid function $\bm{\sigma}(\cdot)$ to map local perception into the probability domain:
\begin{equation}
P_a = 
\begin{cases} 
\bm{\sigma}[\mathbf{W_0^{\top}} \mathbf{x_0}+b_0] & \exists\text{ obstacle}\\
0.5 & \text{other cases}
\end{cases}
,
\label{eq:P_s}
\end{equation}
with open space areas $\mathbf{x_0}$ from the reconstructed local $\Sigma_a$, and the weight vector $\mathbf{W_0}$ in the four quadrants. The bias entry $b_0$ is usually set to zero without a priori. The values of $\mathbf{W_0}$ assign weights to observations in both the navigation scope and the peripheral vision. In our experiments, they are equally weighted with odd symmetry left and right.

\subsubsection{Interaction fields} 
\begin{figure*}[th]
  \centering
  \includegraphics[width=\linewidth,keepaspectratio]{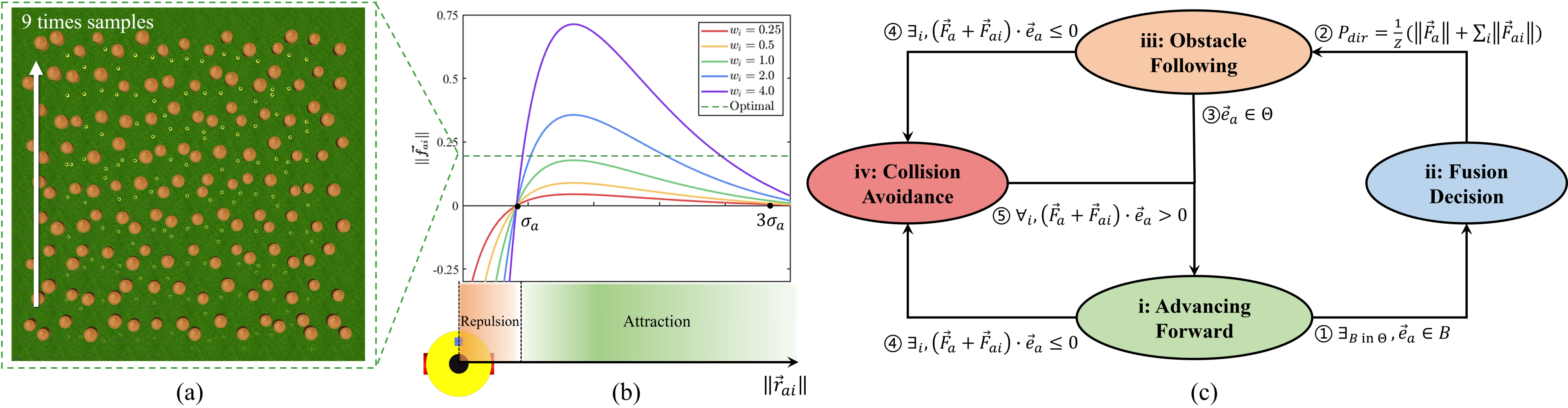}
  \vspace{-0.5cm}
  \caption{(a) Sampling process to determine the optimal $w_i$. (b) Magnitude curves of the interaction fields. (c) Finite state machine and its state transitions.}
  \label{fig:SF}
  \vspace{-0.4cm}
\end{figure*}

They are connections among robots that deliver information flow, regulate swarm coordination, and achieve path optimization. Robots observe their neighbors within the FOV in an ad-hoc way, so we define the interaction field only exerted from local observed neighbors. In Nav-SCOPE, robot connectivity is time-varying according to the visibility and distances of neighboring robots.
Consider a robot $R_a$ with its neighbor $R_i$. 
Consequently, the influence of neighbor \(R_i\) is closely related to its observation features $\Sigma_i$ and their relative distances \( \|\vec{r}_{ai}\| \). The potential \( V_{ai} \) of the interaction combines the effect of coordination convergence $V^{att}_{ai}$ and collision avoidance $V^{rep}_{ai}$ \cite{Social1995}:
\begin{equation}
V_{ai}\left(\left\|\vec{r}_{ai} \right\| \right) =  V^{att}_{ai}\left(\left\|\vec{r}_{ai} \right\| \right) + V^{rep}_{ai}\left(\left\|\vec{r}_{ai} \right\| \right).
\label{eq:neighboring potential}
\end{equation}
To achieve coordination over a moderate distance and implement avoidance when robots are in close proximity, we model the attractive potential with an exponential kernel and the repulsive potential with an inverse proportional kernel:
\begin{equation}
V^{att}_{ai}\left(\left\|\vec{r}_{ai} \right\| \right) = V_0 e^{-\frac{\left\|\vec{r}_{ai} \right\|}{\sigma_a}}, \\
V^{rep}_{ai}\left(\left\|\vec{r}_{ai} \right\| \right) = - \frac{\sigma_a}{\left\|\vec{r}_{ai} \right\|},
\label{eq:V^att_ai}
\end{equation}
where \( \sigma_a \) is the reference distance, marking the transition from repulsion to attraction. The force field $\vec{f}_{ai}$ is the negative gradient of the potential with a weight of $w_i$. When \( \|\vec{r}_{ai} \| = \sigma_a \), \( \vec{f}_{ai} = \textbf{0} \), so the formulation of interaction fields is as follows:
\begin{equation}
\begin{gathered}
V_{ai}\left(\left\|\vec{r}_{ai} \right\| \right) =  e^{1-\frac{\left\|\vec{r}_{ai} \right\|}{\sigma_a}} - \frac{\sigma_a}{\left\|\vec{r}_{ai} \right\|}, \\
\vec{f}_{ai} \left(\left\|\vec{r}_{ai} \right\| \right) = - w_i\nabla_{\vec{e}_{ai}} V_{ai} = \frac{w_i}{\sigma_a} \left( e^{1-\frac{\|\vec{r}_{ai} \|}{\sigma_a}} - \frac{\sigma^2_a}{{\|\vec{r}_{ai}\|}^2}      \right) \vec{e}_{ai},
\end{gathered}
\label{eq:f_ai}
\end{equation}
where $\vec{e}_{ai}$ is a unit vector from $R_a$ to $R_i$. The amplitude of $\| \vec{f}_{ai} \|$ is depicted in Fig. \ref{fig:SF}(b). With an increase in $w_i$, the impact of neighbors becomes more pronounced. The optimal value of $w_i$ can be obtained in Section \ref{subsec:parameter determination}, and its robustness is demonstrated in simulations. Furthermore, when $0<\|\vec{r}_{ai} \| \leq \sigma_a$, the interaction force behaves as a repulsion to avoid path interference. When $\sigma_a<\|\vec{r}_{ai}\| \leq 3\sigma_a$, the force is gravitational for cooperative perception. When $ \|\vec{r}_{ai} \| > 3\sigma_a$, the force decays almost to 0. We define $\|\vec{r}_{ai} \|\in(\sigma_a, 3\sigma_a]$ as the pass band for observation fusion. Neighboring robots outside the pass band are either too close or too far away. A segmentation function \(\vec{F}_{ai}\) is designed to perform differently at varying distances:
\begin{equation}
\vec{F}_{ai} = 
\begin{cases} 
\vec{f}_{ai} & \|\vec{r}_{ai} \|\in(0, \sigma_a]\\
P_{ni} \vec{f}_{ai} & \|\vec{r}_{ai} \|\in(\sigma_a, 3\sigma_a]\\
\textbf{0} & \|\vec{r}_{ai} \|\in(3\sigma_a, +\infty)
\end{cases} ,
\label{eq:F_ai}
\end{equation}
where \( P_{ni} \) is the probability of following the neighbor in future paths. \( P_{ni} \) will be high if the neighbor $R_i$ observes few obstacles towards the target of $R_a$. This will result in a strong attractive effect, thus increasing the likelihood of \(R_a\) to follow it. In contrast, if the location of \(R_i\) is blocked by obstacles, \( P_{ni} \) will be lower, leading to a divergence effect. In such cases, \(R_a\) will be more likely to take a different path, which avoids future detours that would otherwise be caused by its observational limitation. 
The definition of \( P_{ni} \) is differentiated by the quadrant in which the neighbor is observed, in accordance with left searching in (\ref{eq:P_s}):
\begin{equation}
P_{ni} = 
\begin{cases} 
\bm{\sigma}[\mathbf{1} ^ {\top} \cdot (\mathbf{x}_{ni}+\mathbf{w}_n)] & R_i \in \{Q_1, Q_3 \}\\
1-\bm{\sigma}[\mathbf{1} ^ {\top} \cdot (\mathbf{x}_{ni}+\mathbf{w}_n)] & R_i \in \{Q_2, Q_4 \}
\end{cases}
,
\label{eq:P_ni}
\end{equation}
where $\mathbf{x}_{ni}$ indicate the estimates of the passable distances towards the target, and $\mathbf{w}_n$ are the distance thresholds of following. They are both 2D vectors corresponding to the scope quadrants ($Q_{1}$ and $Q_{2}$), after receiving and calibrating the neighboring perception $\Sigma_i$.

\subsubsection{Fusion optimization} The interaction fields provide additional information to identify more efficient paths with fewer detours. When \(R_a\) encounters an obstacle, it combines its own intelligence $\vec{F}_{a}$ and the resultant of $\vec{F}_{ai}$ for subsequent planning directions:
\begin{equation}
\begin{aligned}
P_{dir} &= \frac{1}{Z} \left(\|\vec{F}_{a} \| + \sum\nolimits_{\vec{F}_{ai} \cdot \vec{e}_{ai} > 0} \|\vec{F}_{ai} \|    \right), \\
Z &= 1+\sum\nolimits_{\vec{f}_{ai} \cdot \vec{e}_{ai} > 0} (\|\vec{f}_{ai} \|),
\end{aligned}
\label{eq:P_left}
\end{equation}
where $dir=\{\text{left, right}\}$ for ground robots, \(P_{dir}\) refers to the probability of exploration to the left, and $Z$ is a normalization factor.
An illustrative example can be found in Fig. \ref{fig:detail-result}. The decision is based on the principle of Maximum A Posteriori (MAP) to address the subtle slope of the sigmoid function. The fusion process expands the field of view of individual robots, thus significantly reducing the uncertainty of environmental observations. Consequently, redundant paths are eliminated by convergence or divergence effects in terms of robot paths.

\section{Model Deployment}
\label{sec:parameter determination and model deployment}
This section details the necessary steps for the deployment of Nav-SCOPE, including the planning process, the mutual positioning of swarm robots, and the determination of parameters.

\begin{algorithm}[ht]
\setstretch{1.1}
\caption{Nav-SCOPE}
\label{alg:NavSCOPE}
\begin{algorithmic}[1]
\STATE Initialize neighbor indicators ${\mathcal{N}_a}, \mathcal{R}_a \leftarrow \varnothing$
\STATE $\mathcal{S}_a \leftarrow $ CurrentRobotState() in Fig. \ref{fig:SF}(c)

/* Identifying safe directions */
\STATE FFT points: $N_f\leftarrow 2^{\lceil \log_2 M \rceil}$
\STATE $\widetilde s_a(n) \leftarrow $ Truncate $s_a(n)$ within $[0, l_{th}]$
\STATE $y_{a1}(m) \leftarrow \mathscr{F}^{-1}_{N_f}\left[\mathscr{F}_{N_f}[\widetilde s_a(n)] \odot H_1 \right]R_M(m) $
\STATE $\Theta_s \leftarrow \{\theta_m|y_{a1}(m) \geq l_{th}\}$

/* Updating neighboring robots */
\FOR{$R_i$ in swarm robots}
    \IF{Received feature $\Sigma_a$ request}
        \STATE $\mathcal{R}_a$.add($R_i$)
    \ENDIF
    \IF{$R_i$ is observed within the FOV}
        \STATE $\mathcal{N}_a$.add($R_i$)
        \STATE Obtain $\vec{r}_{ai}$ and calculate $\vec{f}_{ai}$ from (\ref{eq:f_ai})
    \ENDIF
\ENDFOR   

/* Event-triggered information flow */
\IF {$\mathcal{S}_a == \text{ii}$}
    \STATE Request and receive features $\Sigma_i$ from $R_i \in \mathcal{N}_a$
    \STATE Calculate $P_{ni}$ from (\ref{eq:P_ni})
\ELSE
    \STATE $P_{ni}=1$
\ENDIF
\IF{$\mathcal{S}_a == \text{ii}$ or $\mathcal{R}_a \neq \varnothing$}
    \STATE $y_{a2}(m) \leftarrow \mathscr{F}^{-1}_{N_f}\left[\mathscr{F}_{N_f}[s_a(n)] \odot H_2 \right]R_M(m) $
    \STATE $\Sigma_a \leftarrow \left \{(\phi_{a,j}, d_{a,j}) | \Delta y_{a2}(j) \Delta y_{a2}(j+1) < 0 \cap \Theta_s \right\}$
    \STATE Transmit $\Sigma_a$ to all the requested robots $R_i \in \mathcal{R}_a$
\ENDIF

/* Force calculation for state transition */
\STATE Calculate $\vec{F}_a, \vec{F}_{ai}$ from (\ref{eq:F_a}) and (\ref{eq:F_ai})
\RETURN $\Theta_s,\vec{F}_a, \vec{F}_{ai}$
\end{algorithmic}
\end{algorithm}

\subsection{Planning Process}
\label{subsec:planning process}
The planning process based on Nav-SCOPE is a minimal reactive solution, which has proved to be versatile in a wide range of applications \cite{Realtime1985, Dynamic1986, Minimal2019, Predictive2021}. A finite state machine (FSM) is illustrated in Fig. \ref{fig:SF}(c), where a robot $R_a$ will go through a series of states $\mathcal{S}_a \in \{\text{i, ii, iii, iv}\}$ planning towards the target. Algorithm \ref{alg:NavSCOPE} illustrates the detailed process of Nav-SCOPE that obtains safe directions and interaction forces for state transition, where $\mathscr{F}_{N_{f}}[\cdot]$ indicates the FFT transform with $N_f$ points, $\mathcal{N}_a$ is the group of observed neighbors, and $\mathcal{R}_a$ contains teammates that ask for environmental features. The information flow is formed only when needed.

During the planning process, the robot $R_a$ needs to update safe direction set (Lines 3-6) and observed neighbor list (Lines 7-15) continuously.
When the target direction is within safe directions, that is, $\vec{e}_{a} \in \Theta_s$, then \(R_a\) will advance straight towards the target ($\mathcal{S}_a=\text{i}$). Otherwise, the target is blocked by an obstacle $B$, and $R_a$ will request a cooperative perception ($\mathcal{S}_a=\text{ii}$, Lines 16-18) combined with its own observation (Lines 22-24) to find an optimal direction $dir$.
A local group will be naturally formed with the neighboring robots $\mathcal{N}_a$, where the environmental features are sent (Lines 8-9 and Line 25 at neighbors) and gathered at $R_a$. Then, the interaction forces are calculated (Line 27), and the probabilistic fusion is achieved by (\ref{eq:P_left}).
Afterwards, $R_a$ will search along the given $dir$ towards a safe way closest to its target:
\begin{equation}
\min_{m \in dir} \{\langle\theta_m; \theta_{\vec{e}_{a}}\rangle \}, \text{ s.t. } \theta_m \in \Theta_s.
\label{eq:F_aB_final}
\end{equation}
Subsequently, $R_a$ will enter the obstacle following state ($\mathcal{S}_a=\text{iii}$). In case there is a dead end, where $\Theta_s = \varnothing$, the robot will rotate in place towards $dir$ while updating $\Theta_s$, before it meets $\Theta_s \neq \varnothing$ and finds a way out. It then keeps a safety margin from the obstacle borders, until it circumnavigates the obstacle, where $\vec{e}_{a} \in \Theta_s$ satisfies again. Finally, it resumes to advance towards the target.
The reactive strategy of Nav-SCOPE diverges from PF methods \cite{Realtime1985, Predictive2021}, as robot paths are generated directly within safe regions. This feature gets rid of complicated computing for repulsive fields, and avoids saddle points commonly encountered in current methods. Compared with Bug algorithms \cite{Dynamic1986, new2012}, Nav-SCOPE introduces cooperative perception and fusion optimization, which achieve much better paths and coordination.

In order to prevent path conflicts among robots, our model incorporates a collision avoidance state ($\mathcal{S}_a=\text{iv}$). When a robot is subjected to any repulsive interaction $\vec{F}_{ai}$ from neighbors greater than its own propulsion, it then enters the collision avoidance state:
\begin{equation}
\exists_i, (\vec{F}_{a}+\vec{F}_{ai}) \cdot \vec{e}_{a} \leq 0.
\label{eq:collision avoidance}
\end{equation}
In such circumstances, robot \(R_a\) will wait for its neighbor to pass until the above condition is no longer satisfied. 

\subsection{Mutual Positioning}
\label{subsec:mutual positioning}
In order to fuse cooperative perception within the same coordinate system, as well as calculate interaction forces, it is essential to determine relative poses and distances between neighboring robots. Instead of using external equipment, an engineering improvement is introduced to achieve mutual positioning in a distributed way. In this study, each robot is equipped with an ultra-wide band (UWB) beacon and four $90^\circ$ directional array antennas with three units, as shown in Fig. \ref{fig:detail-result}(d). The range and azimuth of neighboring robots are decided by the angle of arrival (AOA) of the UWB beacons and the time difference of arrival (TDOA) between antenna units \cite{Standard2020}. The relative poses of robots can be determined from geomagnetism. In real-world experiments, these are achieved by Nooploop NIUB01 IOT models and CMP10A electronic compasses. 

\subsection{Parameter and Variable Determination}
\label{subsec:parameter determination}
The proposed Nav-SCOPE is interpretable, as its parameters are of explicit physical significance. We provide a way to assign parameters and variables without gaps for deployment.
\subsubsection{Safety margin \texorpdfstring{$r$}{r}} It is determined by robot's max velocity \(v_m\), max acceleration \(a_m\), and the sensor delay \(t_d\):
\begin{equation}
r = r_c + v_m t_d + \frac{v_m^2}{2a_m},
\label{eq:r}
\end{equation}
where the first term refers to the radius of robot's circumcircle, the second term represents the distance $s_d$ caused by the sensor delay, and the third term denotes the braking distance $s_b$.

\subsubsection{Reference distance \texorpdfstring{$\sigma_a$}{sigmaa}} It is set to keep the distances between robots greater than $2r$ as follows:
\begin{equation}
\sigma_a = 2(r+s_d+s_b).
\label{eq:sigma_a}
\end{equation}

\subsubsection{Distance threshold \texorpdfstring{$\mathbf{w}_n$}{wn}} It distinguishes passable ways in neighbors' locations, where robots are expected to advance safely for a while towards the target. Hence, it is assigned as twice the total length of the protective model:
\begin{equation}
\mathbf{w}_{n,k} = 2(r+l_{th})+s_d+s_b, k=1,2,
\label{eq:wn}
\end{equation}
where $k$ is the quadrant index. The value of $\mathbf{w}_n$ should be normalized to the same scale as the sensory data $s_a(n)$.

\subsubsection{Interaction force weight \texorpdfstring{$w_i$}{wi}} It modulates the interaction amplitudes in $\vec{f}_{ai}$. The optimal value can be theoretically determined by AdaBoost \cite{Decision1997} with error rates $\epsilon_s$ and $\epsilon_n$ of the corresponding probabilities $P_a$ and $P_{ni}$. If these parameters are not available, they can be estimated with robust performance using extremely small samples. A sampling playground is randomly generated separate from the maps in the simulation, and nine samples of the swarm are used for an approximate estimation. This process is illustrated in Fig. \ref{fig:SF}(a), and the estimated performances are shown in Fig. \ref{fig:sta-result}(b). Consequently, we get $\epsilon_s=0.245$ and $\epsilon_n=0.446$, and the optimal amplitude of $\vec{f}_{ai}$ in (\ref{eq:P_left}) should be:
\begin{equation}
\|\vec{f}_{ai} \| = \frac{\frac{1}{2} \log(\frac{1-\epsilon_n}{\epsilon_n})}{\frac{1}{2} \log(\frac{1-\epsilon_s}{\epsilon_s})} = 0.195.
\label{eq:COND_w_i}
\end{equation}
The value of $w_i$ should be adjusted to approach the optimal value. We set $w_i=1$ in experimental applications and use $w_i=0.25,0.5,2.0,4.0$ in simulations for comparison.

\subsubsection{Self observation \texorpdfstring{$\mathbf{x_0}$}{x0}} This vector records the estimated open space area from $Q_1$ to $Q_4$. Incorporating environmental features and linear interpolation described in Section \ref{subsec:information flow from cooperative perception}, the elements of \(\mathbf{x_0}\) correspond to the reconstructed area with each quadrant:
\begin{equation}
\begin{aligned}
x_{0,k} &=\frac{1}{2}\int^{\phi_{k,u}}_{\phi_{k,l}} \left(\frac{d_{k,u} - d_{k,l}}{\phi_{k,u}-\phi_{k,l}} \phi+ \frac{d_{k,l}\phi_{k,u} - d_{k,u}\phi_{k,l}}{\phi_{k,u}-\phi_{k,l}}  \right)^2 d\phi \\
&= \frac{\phi_{k,u}-\phi_{k,l}}{6} \left(d_{k,l}^2 + d_{k,l}d_{k,u} + d_{k,u}^2  \right)
, k=1,\cdots,4
\end{aligned}
\label{eq:x_i}
\end{equation}
where $\phi_{k,u}$ and $\phi_{k,l}$ are the upper and lower boundaries of the \( k \)-th quadrant, and $d_{k,u},d_{k,l}$ are the corresponding distances.

\subsubsection{Neighbor observation \texorpdfstring{$\mathbf{x}_{ni}$}{xni}} Its elements are estimates of passable distances described in (\ref{eq:P_ni}). In order to identify open spaces with fewer obstacles, \(\mathbf{x}_{ni}\) is assigned in a fuzzy manner:
\begin{equation}
x_{ni,k} = 
\begin{cases} 
\left(\max{d_{i,k}}, \min{d_{i,k}}  \right) /2 & \exists\text{ local minima}\\
\max{d_{i,k}} & \text{other cases}\\
\end{cases}
,
\label{eq:x_1}
\end{equation}
where $k=1,2$ for $Q_1$ and $Q_2$. Since local minima indicate potential obstacles, the minimum value delivers a message of potential obstruction. For the case of no local minima, the maxima value can effectively emphasize potential pathways. 

%% file: Simulation_Results.tex
\section{Simulation Results}
\label{sec:simulation results}
\begin{figure*}[ht]
  \centering  \includegraphics[width=\linewidth,keepaspectratio]{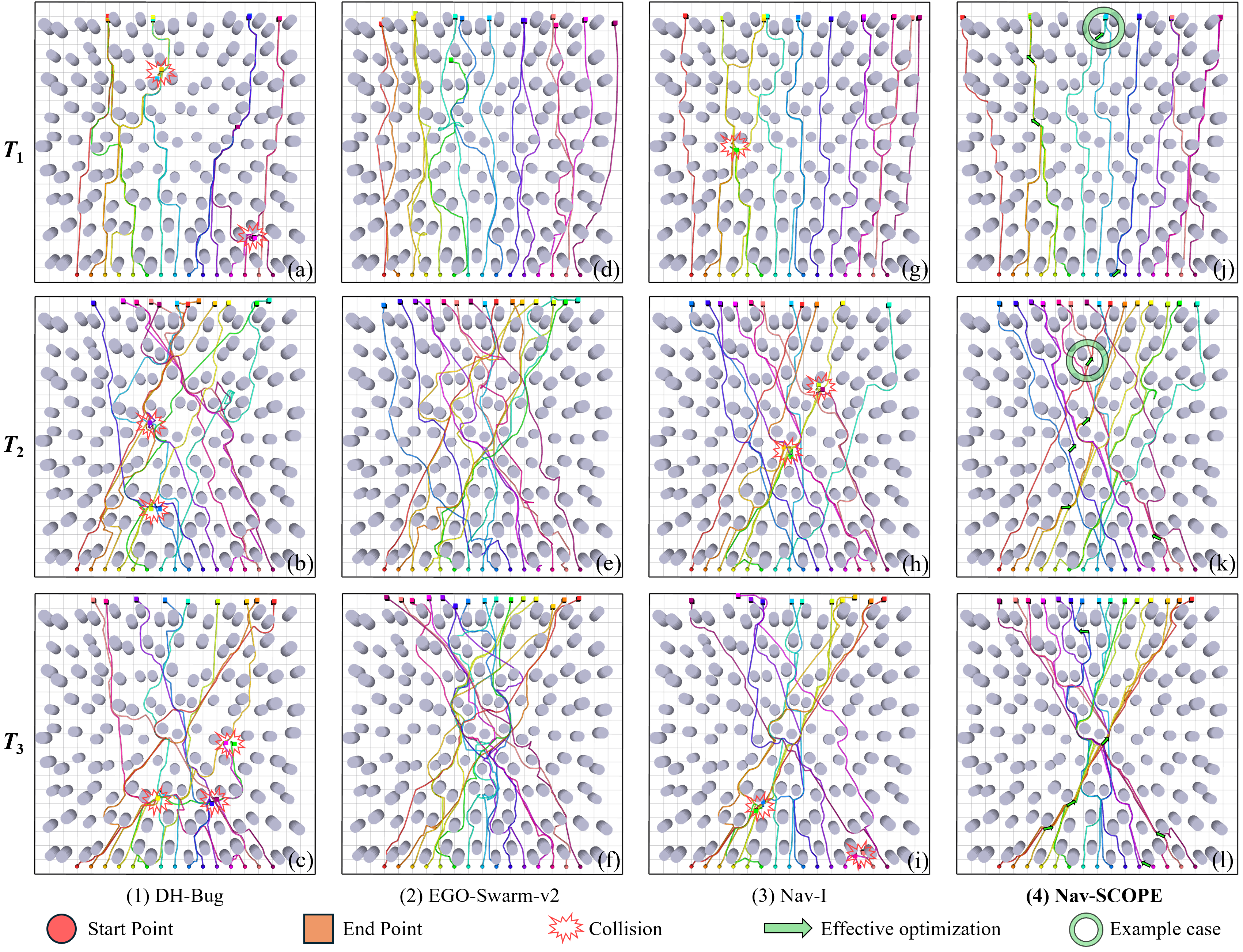}
  \vspace{-0.6cm}
  \caption{Result comparison between Nav-SCOPE and baselines. Each row corresponds to a specific task, and the columns represent four algorithms.}
  \label{fig:traj-result}
  \vspace{-0.4cm}
\end{figure*}
In this section, Nav-SCOPE is applied to robot swarms in cluttered and unknown environments on Gazebo using an i9-12900K CPU. We compare the performance of Nav-SCOPE with 3 baselines: a current reactive solution DH-Bug \cite{new2012}, a current SOTA method EGO-Swarm-v2 \cite{Swarm2022}, and an ablation baseline $w_i = 0$ referred to as Nav-I, concerning individual robots without neighboring scope to form the interaction fields. 

To evaluate performances, we use the following statistical metrics: (a) Robot arrival rate (AR), indicating the ratio of arrived robots to the total number of swarm robots. (b) Path redundancy $\gamma$, referring to the redundant path relative to the optimal path length. Inspired from the information theory \cite{Shannon1948}, $\gamma$ is defined as
\begin{equation}
    \gamma = 1- AR \cdot \frac{\Sigma_i l_i^*}{\Sigma_i\bar{l}_i},
\end{equation}
where $i$ is the robot index within a swarm, $l^*$ signifies the optimal path length, and $\bar{l}$ denotes the average path length. Here, we use the path from A* \cite{Formal1968} with a resolution of 0.05m as the optimal path. (c) Duration of the task $t$.

Swarms of 15 robots are tested in very dense forests, about 0.35 obstacles$/m^2$ density in a 400$m^2$ region. The physical characters of the robot are $r_0=$ 0.15m, $v_m=$0.5m/s, and $a_m=$ 2$\text{m}/\text{s}^{2}$. The sensor we used is a 5Hz LiDAR with an observation range of $R=$5m. The parameters are: $r=$ 0.30m, $l_{th}=$ 0.60m, $\alpha=$60$^{\circ}$, and $\sigma_a=$ 0.9m. The maximum simulation time is set to 5 minutes. We compared performance in three tasks: 
\begin{itemize}
    \item $T_1$: Cooperative traversing from one side to the other.
    \item $T_2$: Intersected targets with different start and end points.
    \item $T_3$: Enforced collision avoidance with extreme interferences.
\end{itemize}

The statistical results for 20 random maps are detailed in Table \ref{sta-result} and Fig. \ref{fig:sta-result}(a). The optimal parameter $w_i$ aligns with the sampled results in Section \ref{subsec:parameter determination}, as illustrated in Fig. \ref{fig:sta-result}(b). 

\begin{table}[th]
\small
\caption{Statistical results of Nav-SCOPE and baselines}
\label{sta-result}
\begin{center}
\renewcommand{\arraystretch}{1.15} 
\begin{tabular}{c c c c c}
\toprule
Task & Algorithm & $\gamma$ (\%) & AR (\%)& $t$ (s)\\
\midrule
\multirow{3}{*}{$T_1$} & DH-Bug & 16.3& 90.3& 88.0\\
 & EGO-Swarm-v2& 10.6& 94.7& 123.4\\
 & Nav-I& 13.1& 92.7& 81.8\\
 & \textbf{Nav-SCOPE} & \textbf{5.1}& \textbf{100}& \textbf{66.2}\\
 \hline
\multirow{3}{*}{$T_2$} & DH-Bug & 27.6& 79.7& 125.5\\
 & EGO-Swarm-v2& 15.5& 92.7& 149.4\\
 & Nav-I& 30.2& 75.7& 132.9\\
 & \textbf{Nav-SCOPE} & \textbf{4.4}& \textbf{99.7}& \textbf{85.0}\\
 \hline
\multirow{3}{*}{$T_3$} & DH-Bug & 34.1& 75.0& 140.4\\
 & EGO-Swarm-v2& 24.7& 85.3& 184.5\\
 & Nav-I& 35.8& 71.3& 145.2\\
 & \textbf{Nav-SCOPE} & \textbf{7.1}& \textbf{98.7}& \textbf{99.0}\\
\bottomrule
\end{tabular}
\end{center}
\vspace{-0.4cm}
\end{table}

Fig. \ref{fig:traj-result} depicts results from a random forest in three scenarios. Many redundant paths can be observed in three baselines, mainly caused by limited FOV, obstacle obstructions, and robot intersections. DH-Bug performs greedy reactive behaviors towards the target. But sometimes more haste, less speed. It often chooses worse directions with more detours and conflicts. EGO-Swarm-v2 impressively identifies local open spaces to generate long-term paths with much fewer circumnavigations. However, the restricted observation occasionally results in deviations and reversals. Nav-I performs suboptimal navigation without interaction fields from neighbors, making it almost degenerate into DH-Bug. In contrast, the performance of Nav-SCOPE is robust across different tasks. Coordinated navigation is achieved with the information flow from cooperative perception. Specifically, in the following aspects:

\begin{figure}[ht]
  \centering  \includegraphics[width=\linewidth,keepaspectratio]{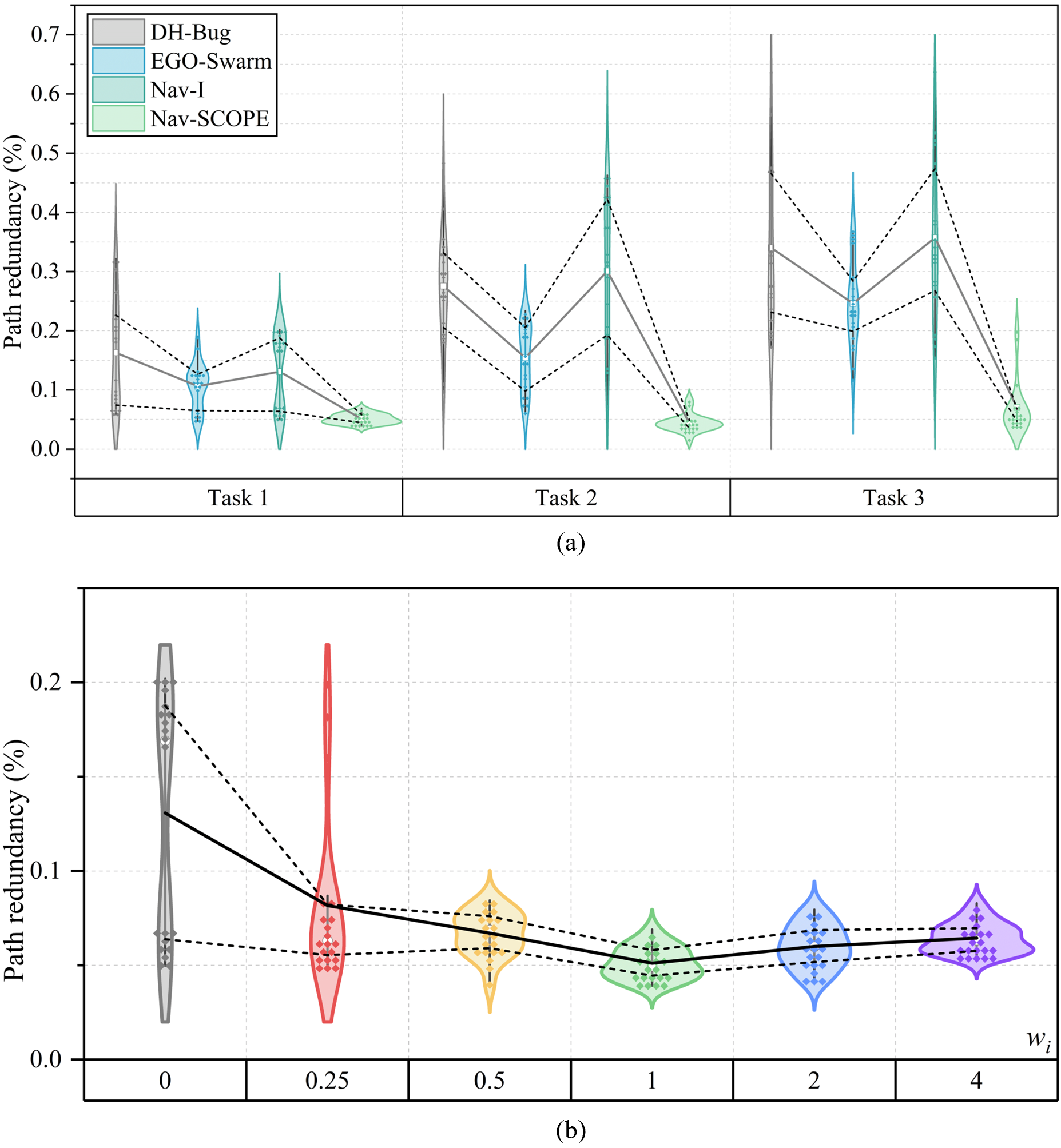}
  \vspace{-0.5cm}
  \caption{(a) Path redundancy $\gamma$ across different tasks and algorithms. (b) Path redundancy $\gamma$ with different $w_i$.}
  \label{fig:sta-result}
  \vspace{-0.4cm}
\end{figure}

\subsection{Convergence effects} The arrow markings in Fig. \ref{fig:traj-result}(j)-(l) show some cases where paths are optimized through spontaneous synergies within local teams. A detailed case is depicted in Fig. \ref{fig:detail-result}(a). When a robot identifies a safe gap with fewer obstacles, it attracts nearby neighbor robots to follow its path. As a result, local teams are naturally formed by convergence effects along paths with relatively fewer obstacles. This type of effect is absent from the baseline methods. Specifically, the ablation study of Nav-I indicates that optimization with local perception alone has limited impacts on long-term performance, particularly in complicated tasks.

\subsection{Divergence effects} Note that the information from the cooperative perception may be negative when neighbors detect obstacles. Fig. \ref{fig:detail-result}(b) illustrates a case when the robot $R_b$ in front encountered an obstacle, the following robot $R_a$ turned to the other side. Thus, robots can avoid unnecessary detours caused by perceptual limitations and optimize future paths with fused observation. EGO-Swarm-v2 also exhibited divergence effects as it tends to generate scattered paths to avoid collisions. But this function sometimes leads to unnecessary detours.

\subsection{Collision avoidance} This feature is achieved by the repulsive effect of the interaction field, where planning directions are considered to prevent future interference between neighboring robots. Simulation results demonstrate safe and organized paths, especially when robots are enforced to avoid reciprocal collisions, as shown in Fig. \ref{fig:traj-result}(l). However, for EGO-Swarm-v2, the loss of DOF on the Z axis makes it challenging to find a collision-free path for mobile robots in dense situations. Hence, planner timeouts and aggressive trajectories occasionally occurred, resulting in robot failures and longer task durations.

\begin{table}[htbp]
  \small
  \centering
  \caption{computation and communication usage}
  \label{cnc_results}
  \renewcommand{\arraystretch}{1.5} 
  \begin{tabularx}{1.0\linewidth}{ m{1.75cm}<{\centering} m{1.1cm}<{\centering}  m{1.1cm}<{\centering}  m{1.3cm}<{\centering} m{1.7cm}<{\centering}}
    \hline
      & $t_{\text{plan}}$ ($\upmu$s) & Memory (KB) & Bandwidth$^1$ (Kbps) & Package Size (Byte)\\
    \hline
    DH-Bug & 1.8 & 1.6 & / & /\\
    Nav-I & 15.8 & 14.8 & / & /\\
    EGO-Swarm-v2 & 1.2$\times$10$^3$ & 3.5$\times$10$^3$ & 316.5 & 289.9\\
    Nav-SCOPE & 15.9 & 14.9 & 33.6 & 80.8\\
    \hline
  \end{tabularx}
  \begin{tablenotes}
     \item[1] $^1$ Syetem level.
   \end{tablenotes}
   \vspace{-0.5cm}
\end{table}

\subsection{Resource consumption} The statistical performances of computation and communication are summarized in Table \ref{cnc_results}, where $t_{\text{plan}}$ denotes the end-to-end time delay from perception to planning execution. In terms of computation, Nav-SCOPE performs the same order as minimal reactive methods and is much lower compared to EGO-Swarm-v2. The computation mainly comes from distributed FFT filtering and feature extraction on each robot.
Compared with Nav-I, we can see that the calculation of interaction fields costs only 0.1 $\upmu$s on average. This is achieved by our ad-hoc connectivity design, which reduces the complexity of scaling to linear order.
As for communication, Nav-SCOPE consumes a minimal bandwidth while carrying more information from cooperative perception. This is realized by compressed features in (\ref{eq:extrema}) and on-demand transmission. Consequently, Nav-SCOPE is an ultralight solution, suitable for deployment on robots with limited computational capacity or in scenarios with restricted communication.

%% file: Real_World_Experiments.tex
\section{Real World Experiments}
\label{sec:real world experiments}
Besides simulations, we also performed real-world experiments with a swarm of 5 robots. Each robot is equipped with an 8-core ARM v8.2 CPU and a 20Hz OS1 LiDAR with an effective range of 5m. Data transmission is performed by a 78.125 Kbps ad-hoc UWB network. The robots use Nooploop NIUB01 IOT models and CMP10A electronic compasses for mutual positioning. The robot characters are:
$r_0=$ 0.29m, $r_e=$ 0.43m, $v_m=$ 0.5m/s, $a_m=$ 4m/s$^2$. Consequently, we set $r=$ 0.49m, $l_{th}=$ 0.83m, $\alpha=$ 72$^{\circ}$ and $\sigma_a=$ 1.09m. The LiDAR observation is represented as a polar coordinate array with approximately 360 points.

\begin{figure}[th]
  \centering  \includegraphics[width=\linewidth,keepaspectratio]{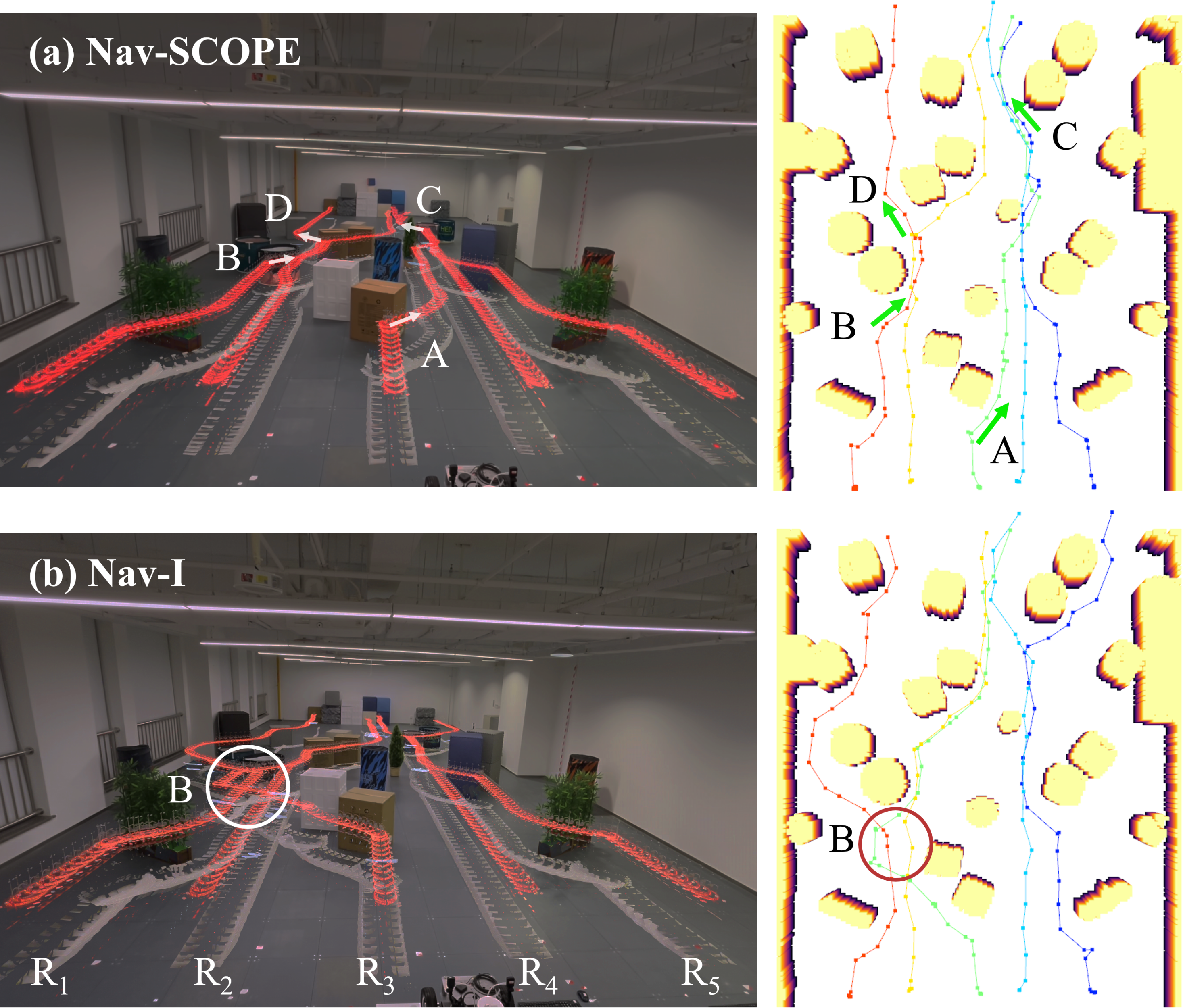}
  \vspace{-0.5cm}
  \caption{Experimental results in a cluttered depository. To present the obstacles and trajectories with clarity, a point cloud map was produced \textit{after} the experiments finished.}
  \label{fig:img-real}
\end{figure}

\begin{figure}[ht]
  \centering
  \includegraphics[width=\linewidth,keepaspectratio]{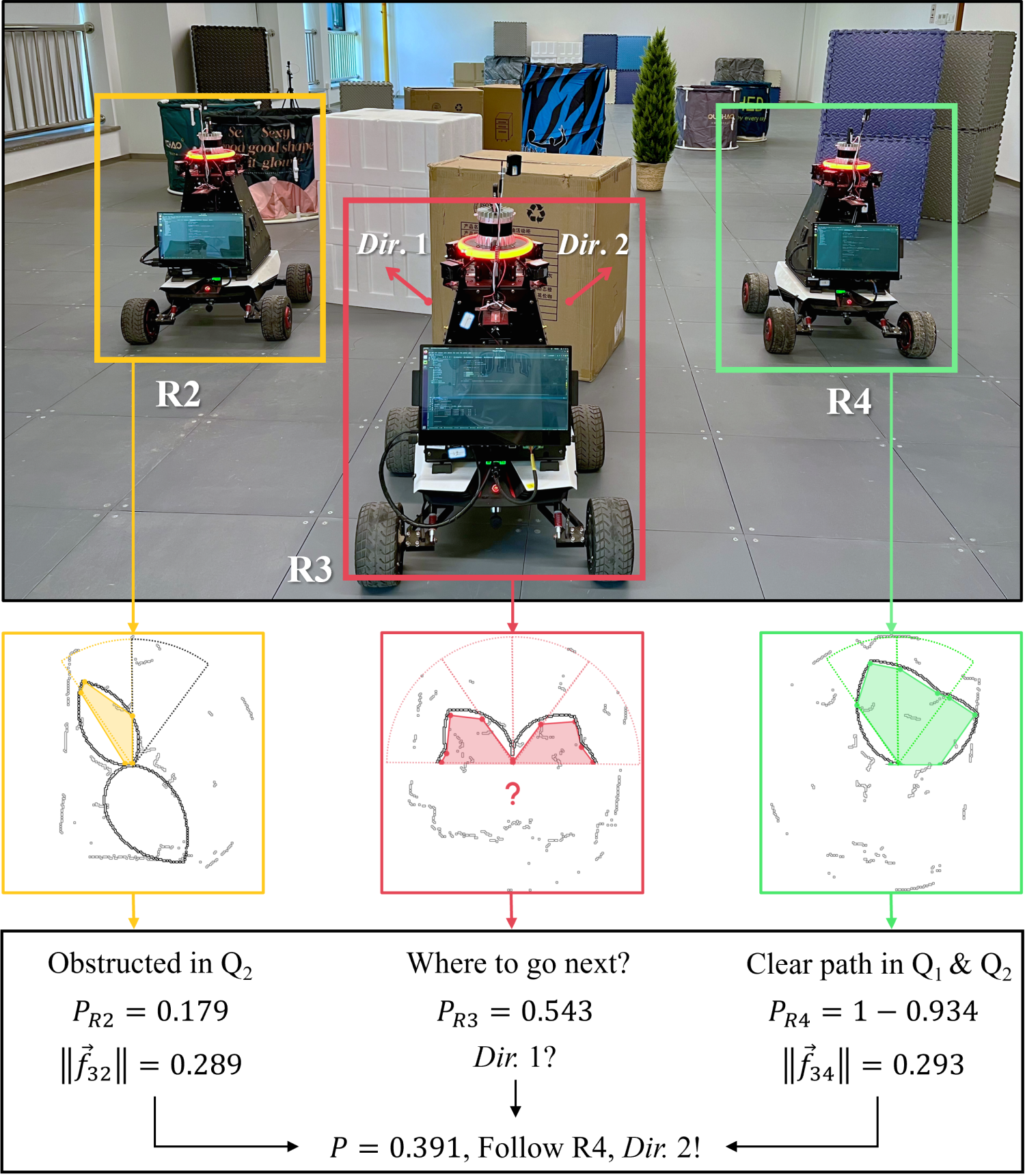}
  \vspace{-0.5cm}
  \caption{An effective case of coordinated navigation within a local team.}
  \label{fig:real-example}
\end{figure}

The experimental setup consists of a cluttered depository containing various shaped obstacles. The density is approximately 0.38 obstacles$/m^2$. The result is illustrated in Fig. \ref{fig:img-real}. Our proposed Nav-SCOPE generates coordinated and collision-free trajectories in contrast to the individual Nav-I. The average path length of Nav-SCOPE is $10.77\text{m}$ with approximate $\gamma=3.8\%$, while Nav-I achieves $11.29\text{m}$ with much higher redundancy $\gamma=8.3\%$.

Specifically, we observe convergence effects at points $A$, $B$ and $C$, and a divergence effect at point $D$. All of these are accomplished by incorporating supplementary information from cooperative perception. A particular case at point $A$ is presented in Figure \ref{fig:real-example}, where it is difficult for $R_3$ to decide its future direction, as the current observation looks similar on either side. But after fusion, it chose to follow $R_4$ rightwards, and there proved to be fewer obstacles. The swarm also gains advantages from collision avoidance by eliminating path interferences which evidently occurred at points $D$ in the Nav-I baseline. In contrast, Nav-SCOPE shows organized trajectories despite robot interactions and narrow pathways.

%% file: Conclusion.tex
\section{Discussion and Conclusion}
\label{sec:Conclusion}

Despite the current success, we acknowledge that Nav-SCOPE is only applicable to ground robots at this stage, while extension is considered for applications in 3D space. Our future aim is to implement Nav-SCOPE on aerial robots, such as quadrotors. In the perception stage, we will introduce 2D FFT filtering to process depth images \mbox{\cite{Two-dimensional1975}}. In the planning stage, we will extend the probability function from bi-categories to multi-categories, and incorporate trajectory optimization with robot dynamic costs \mbox{\cite{Computationally2015}}. In execution, we will add the time-delay calibration in high-speed scenarios \mbox{\cite{Consensus2022}}. 

Other promising areas include model adaptation to dynamic environments and task expansion to exploration applications. Some existing theories such as obstacle tracking \mbox{\cite{Leveraging2020}} and target allocation \mbox{\cite{Development2025}}, etc. could be incorporated to improve system performance in specific tasks.

In summary, we propose a systematic solution to achieve cooperative perception and coordinated navigation on swarm robots. Our approach facilitates information flow for observation fusion, followed by path optimization with self-organized convergence, divergence, and collision avoidance. Baseline comparisons demonstrate its exceptional performance, robustness, and lightweight demands.